\documentclass{llncs}  

\title{
User Loss -- A Forced-Choice-Inspired Approach to Train Neural Networks directly by User Interaction
}
\usepackage{authblk}
\usepackage{graphicx}
\usepackage{tikz}
\usepackage{multirow}
\usepackage{amssymb}
\usepackage{amsmath}
\usepackage{float}
\usepackage{subcaption}
\usetikzlibrary{shapes.arrows,chains,positioning}

\author{Shahab Zarei, Bernhard Stimpel, Christopher Syben, Andreas Maier} 
\institute{Pattern Recognition Lab, Department of Computer Science,
Friedrich-Alexander-Universit\"at Erlangen-N\"urnberg, Germany}
\titlerunning{User Loss - A Forced-Choice-Inspired Approach}
	\authorrunning{Zarei et al.}
\begin{document}
\maketitle
\thispagestyle{empty}
\pagestyle{empty}

\begin{abstract}
In this paper, we investigate whether is it possible to train a neural network directly from user inputs. We consider this approach to be highly relevant for applications in which the point of optimality is not well-defined and user-dependent. Our application is medical image denoising which is essential in fluoroscopy imaging. In this field every user, i.e. physician, has a different flavor and image quality needs to be tailored towards each individual.

To address this important problem, we propose to construct a loss function derived from a forced-choice experiment. In order to make the learning problem feasible, we operate in the domain of precision learning, i.e., we inspire the network architecture by traditional signal processing methods in order to reduce the number of trainable parameters. The algorithm that was used for this is a Laplacian pyramid with only six trainable parameters.

In the experimental results, we demonstrate that two image experts who prefer different filter characteristics between sharpness and de-noising can be created using our approach. Also models trained for a specific user perform best on this users test data. This approach opens the way towards implementation of direct user feedback in deep learning and is applicable for a wide range of application.
\end{abstract}

\section{Introduction}

Deep learning is a technology that has been shown to tackle many important problems in image processing and computer vision \cite{lecun2015deep}. However, all training needs a clear reference in order to apply neural network-based techniques. Such a reference can either be a set of classes or a specific desired output in regression problems. However, there are also problems in which no clear reference can be given. An example for this are user preferences in forced-choice experiments. Here, a user can only select the image he likes best, but he cannot describe or generate an optimal image. In this paper, we tackle exactly this problem by introduction of a user loss that can be generated specifically for one user of such a system. 

In order to investigate our new concept, we explore its use on image enhancement of interventional X-ray images. Here, the problem arises that different physicians prefer different image characteristics during their interventions. Some users are distracted by noise and prefer strong de-noising while others prefer crisp and sharp images. Another requirement for our user loss is that we want to spend only few clicks for training. As such we have to deal with the problem of having only few training samples, as we cannot ask your users to click more than 50 to 100 times. In order to still work in the regime of deep learning, we employ a framework coined {\it precision learning} that is able to map known operators and algorithms onto deep learning architectures \cite{precision_learning}. In literature this approach is known to be able to reduce maximal error bounds of the learning problem and to reduce the number of required training samples \cite{filter_learning}. Fu et al. even demonstrated that they are able to map complex algorithms such as the vesselness filter onto a deep network using this technique \cite{vesselnet}.

\section{Methods}
For this paper, we chose an Laplacian pyramid de-noising algorithm as basis \cite{c4}. In this section first image denoising using the Laplacial pyramid is described. Then, we follow the idea of precision learning to derive the network topolgy based on the known approach followed by an detailed description of the loss function.
\subsection{Subband decomposition}
Image densoising using a Laplacian pyramid is carried out in two steps. First the image is decomposed into subbands followed by an soft threshold to reduce the noise.
The Laplacian pyramid \cite{c4} is an extension of the Gaussian pyramid using differences of Gaussians (DoG). To construct a layer of the Laplacian pyramid the input has to be blurred using a Gaussian kernel with a defined standard deviation $\sigma$ and mean $\mu=0$ with a subsequent subtraction from the unblurred input itself. This difference image is one layer in the Laplacian pyramid, while the blurred input image is downsampled by a defined factor $k$ serves as the input for the next layer. Repeating this \textit{Smoothing}, \textit{Subtraction} and \textit{Down-sampling} $n$ times constructs a pymarid of depth $n$. The Gaussian parameters have to be defined for each layer, thus the construction of the pyramid can be described with:
\begin{align}
\vec I_{\text{lp},n} &= \vec G_{\sigma_n} * \vec I_n
\qquad\text{with}\qquad
G(\sigma_n) = \frac{1}{2\sqrt{\sigma_n}}\exp\left(\frac{-x^2}{2|\sigma_n|^2}\right)\\
\vec I_{\text{bp},n} &= \vec I_n - \vec I_{\text{lp},n} \enspace ,
\end{align}
where $\vec I_n$ is the input image for layer $n$, $\vec G_{\sigma_n}$ the Gaussian kernel described by the standard deviation $\sigma_n$ for the respective layer, $\vec I_{\text{lp},n}$ the low-pass image, and $\vec I_{\text{bp},n}$ is the bandpass image which represents the layer of the Laplacian pyramid. 
\color{black}

\subsection{Soft-Thresholding}



\begin{figure}[tb]
	\centering
    \resizebox{\columnwidth}{!}{
	\begin{tabular}{cc}
		\begin{subfigure}{0.41\textwidth}\centering\includegraphics[width=1\columnwidth]{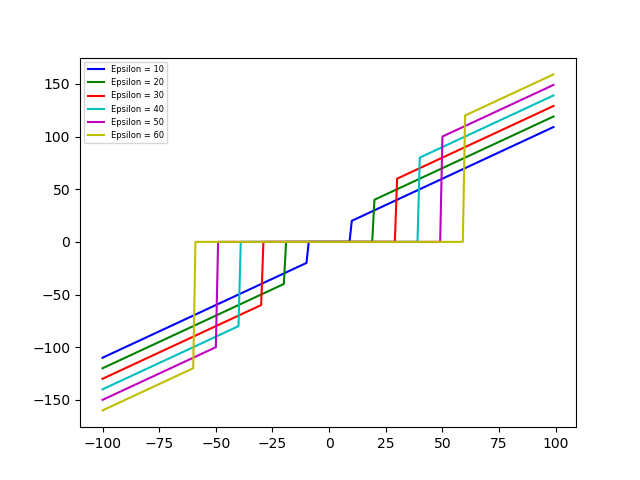}\caption{}\end{subfigure}&
		\begin{subfigure}{0.41\textwidth}\centering\includegraphics[width=0.85\columnwidth]{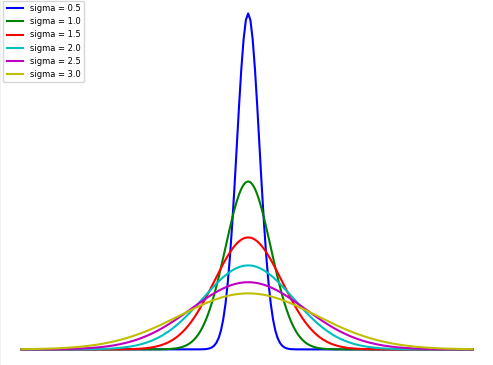}\caption{}\end{subfigure}
	\end{tabular}
    }
	\caption{(a) Soft threshold and (b) Gaussian kernel used in this work.}
	\label{tab:mytable}
\end{figure}

After sub-band decomposition, we assume that small coefficients are caused by noise of different strength in each sub-band $\vec I_{\text{bp},n}$. Here, we employ a soft-thresholding technique to suppress this noise with magnitudes smaller than $\epsilon$:

\begin{equation}
\begin{array}{l}
\text{soft}(x, \epsilon) = 
\begin{cases}
\text{sign}(x)(|x|- \epsilon) & \text{if} \quad |x|\geq\epsilon  \\
0 & \text{otherwise}\\
\end{cases}
\end{array}
\end{equation}
Note that for both, the Gaussian that is used for the sub-band decomposition, as well as for the soft thresholding function sub-gradients \cite{rockafellar} can be computed with respect to their parameters (cf. Fig.~\ref{tab:mytable}). As such both are suited for use in neural networks \cite{precision_learning}.

\subsection{Neural Network}
\begin{figure}[tb]
\centering
\includegraphics[width=\linewidth]{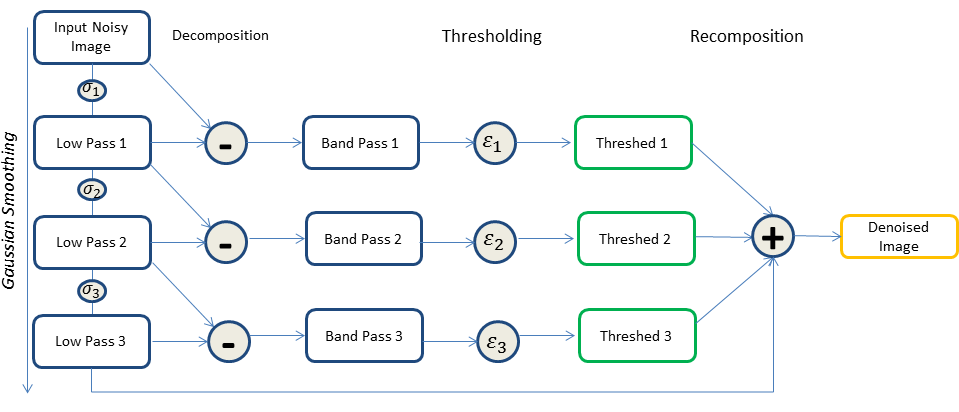}
\caption{Schematic of the neural network design used in this work. The architecture mimics a Laplacian pyramid filter with soft-thresholding}\label{fig:network}
\end{figure}

Following the {\it precision learning} paradigm, we construct a three layer Laplacian pyramid filter as a neural network. A flowchart of the network is depicted in Fig.~\ref{fig:network}. The low-pass filters are implemented as convolutional layers, in which the actual kernel only has a single free parameter $\sigma$. Using point-wise subtraction, these low-pass filters are used to construct the band-pass filters. On each of those filters, soft-thresholding with parameter $\epsilon$ is applies. In a final layer, the soft-thresholded band-pass filters are recombined to form the final image. As such we end up with a network architecture with nine layers that only has six trainable parameters $\sigma_1, \sigma_2, \sigma_3, \epsilon_1, \epsilon_2, \epsilon_3$. In the following, we summarize these parameters as a single vector $\vec{\phi}$ that can be trained using the back-propagation algorithm \cite{rumelhart1986learning}.

\subsection{User Loss}
Let $\vec I_\text{pref}$ be the user preferred image, $\vec I_\text{NN}$ the denoised image produced by our network. Below equation would be the main objective of our NN de-noiser:

\begin{equation}
	\text{argmin}_{\vec \phi}\; ||\vec I_\text{pref} -  \vec I_\text{NN}||_2^2
\end{equation}
The main problem with this equation is that the user is not able to produce $\vec I_\text{pref}$. To resolve this problem, we introduce errors to the optimal image that cannot be observed directly:

$$e = ||\vec I_\text{pref} -  \vec I||_2^2$$

However, if we provide a forced-choice experiment using four images $\vec I_0 \ldots \vec I_0$, we can determine which of the four errors $e_0 \ldots e_3$ is the smallest. 
This gives us a set of constraints that need to be fulfilled by our neural network.
For the training of the network, we define our error in the following way:

$$e_q = ||\vec I_\text{NN} -  \vec I_q||_2^2$$

Let $s$ be the total number of frames, $e_{s,q}$ denote the quality $q$ dedicated to frame $s$, and $Q$ denote the number of choices. Assuming $e_{s,*}$ is selected by the user, the following expected relationships between the errors emerge:

\begin{equation}
\begin{array}{l}
e_{s,*} \leq e_{s,q} \qquad \forall q \in \{0, \ldots, Q-1\} \\
\end{array}
\end{equation}
For user selection is $* = 2$, the constraint below are used to set up our loss function. Similar to implementation of support vector machines in deep networks, we map the inequality constraints to the hinge loss using the $\max$ operator \cite{Bishop:2006}.
\begin{equation}
\begin{array}{l}
    e_{s,2} \textless e_{s,0} \quad \xrightarrow{\text{ }} \quad e_{s,2} - e_{s,0} \textless 0 \quad \xrightarrow{\text{ }} \quad \max(e_{s,2} - e_{s,0}, 0)\\
    e_{s,2} \textless e_{s,1} \quad \xrightarrow{\text{ }} \quad e_{s,2} - e_{s,1} \textless 0 \quad \xrightarrow{\text{ }} \quad \max(e_{s,2} - e_{s,1}, 0)\\
    e_{s,2} \textless e_{s,3} \quad \xrightarrow{\text{ }} \quad e_{s,2} - e_{s,3} \textless 0 \quad \xrightarrow{\text{ }} \quad \max(e_{s,2} - e_{s,3}, 0)    
\end{array}
\end{equation}
This gives rise to three different variants of the user loss that are used in this work: 
\begin{enumerate}
\item Best-Match: Only the user selected image is used to guide the loss function:	
	\begin{equation}
	\text{argmin}_{\vec \phi} \sum_{s=1}^{S} e_{s,*} 
	\end{equation}

\item Forced-Choice: The user loss seeks to fulfill all criteria imposed by the user selection. 
	\begin{equation}
	\text{argmin}_{\vec \phi} \sum_{s=1}^{S} \sum_{q=0}^{Q-1} \max(e_{s,*} - e_{s, q}, 0) 
	\end{equation}
    
\item Hybrid: The user selected image drives the parameter optimization while all constraints implied by the forced-choice are sought to be fulfilled.
	\begin{equation}
	\text{argmin}_{\vec \phi} \sum_{s=1}^{S} e_{s, *} + \sum_{q=0}^{Q-1} \max(e_{s, *} - e_{s, q}, 0)  
	\end{equation}
\end{enumerate}  
Note that the hybrid user loss is mathematically very close to the soft-margin support vector machine, where $e_{s, *}$ takes the role of the normal vector length and $\sum_{q=0}^{Q-1} \max(e_{s, *} - e_{s, q}, 0)$ the role of the additional constraints.

\section{Experiments and Results}

\begin{figure}[tb]
	\begin{center}
		\includegraphics[width=\linewidth]{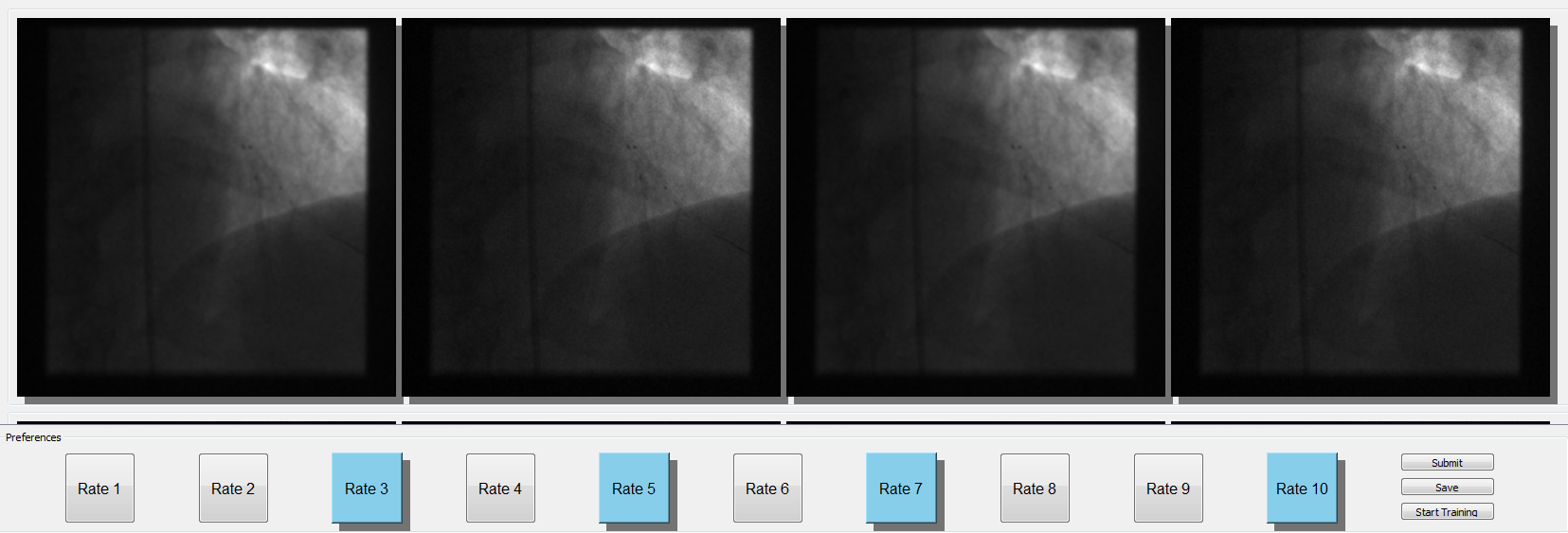}
		\caption{Graphical user interface designed for proposed network training.}\label{fig:gui}
	\end{center}
\end{figure}

For generating different scenarios, in the first step the Laplacian pyramid is initialized for each input image. 
Considering the center values of our parameter sets $\vec \phi$, the four different scenes are generated using random parameters. 
The resulting scenes for each frame are then imported to a GUI in order to take the user preferences (cf. Fig.~\ref{fig:gui}).


The network is implemented in Python using Tensorflow framework. ADAM algorithm is used as optimizer iterating over 5000 epochs with learning rate of $\mu = 10^{-2}$ and the batchsize is set to 50. 

The datasets which are used in this work are 2D angiography fluoroscopy image data.
The dataset contains 50 images of size $1440\times1440$ with different dose levels. We created 200 scenarios via randomly initializing the Laplacian pyramid parameters.
Our dataset is divided such that 60\% of the dataset for training data, 20\% for validation and 20\% for test set. In this work \textit{stratified K-Fold Cross-Validation} is used for data set splitting.

\subsection{Qualitative Results}
\begin{figure}[tb]
	\centering
	\resizebox{\columnwidth}{!}{
		\begin{tabular}{ c c c c}
			Original & Best Match & Forced Choice & Hybrid\\
			
			\begin{subfigure}{0.5\textwidth}\centering\includegraphics[width=0.9\columnwidth]{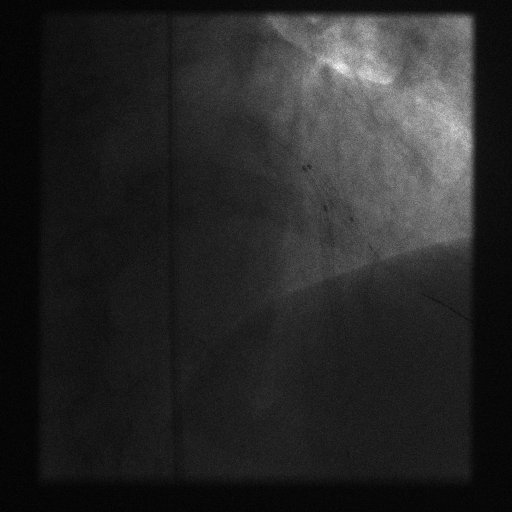}\label{fig:taba}\end{subfigure}&
			\begin{subfigure}{0.5\textwidth}\centering\includegraphics[width=0.9\columnwidth]{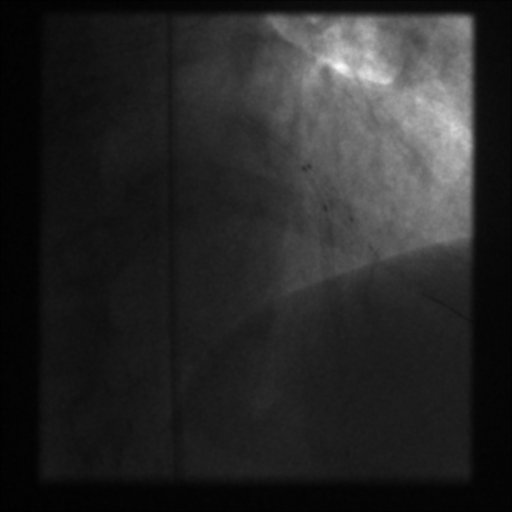}\label{fig:taba}\end{subfigure}&
			\begin{subfigure}{0.5\textwidth}\centering\includegraphics[width=0.9\columnwidth]{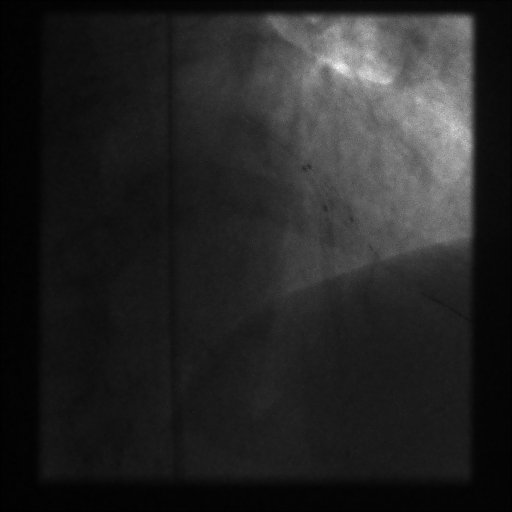}\label{fig:taba}\end{subfigure}&
			\begin{subfigure}{0.5\textwidth}\centering\includegraphics[width=0.9\columnwidth]{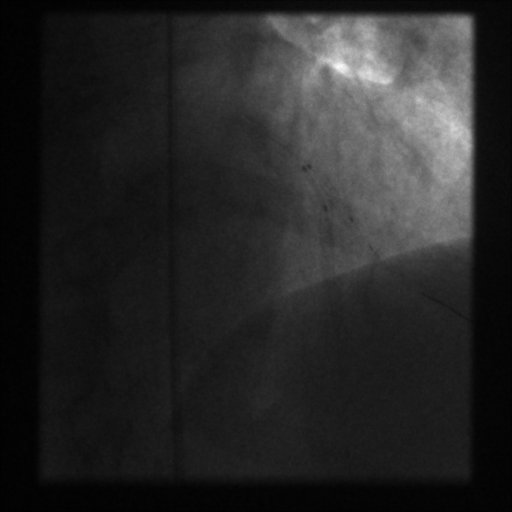}\label{fig:taba}\end{subfigure} \\\\
			
			\begin{subfigure}{0.5\textwidth}\centering\includegraphics[width=0.9\columnwidth]{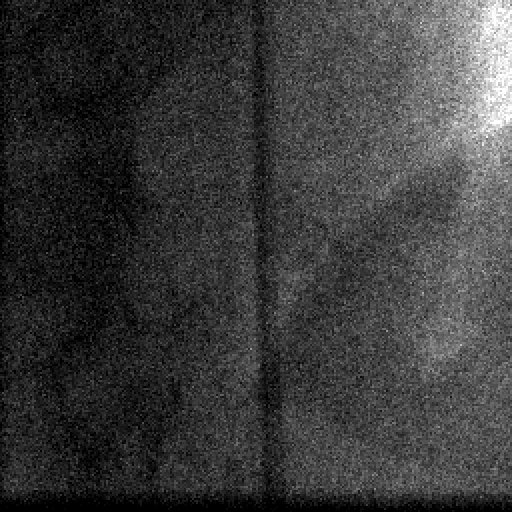}\label{fig:taba}\end{subfigure}&
			\begin{subfigure}{0.5\textwidth}\centering\includegraphics[width=0.9\columnwidth]{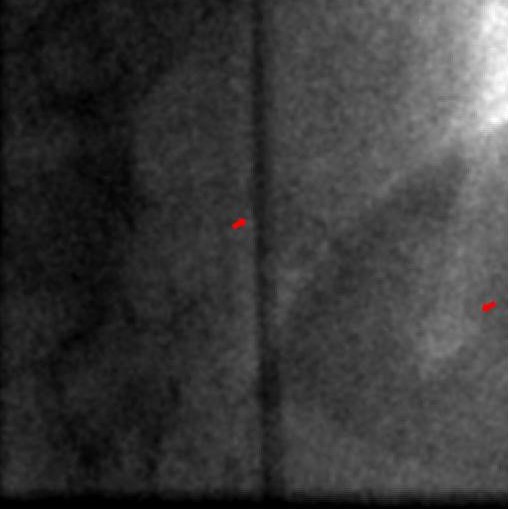}\label{fig:taba}\end{subfigure}&
			\begin{subfigure}{0.5\textwidth}\centering\includegraphics[width=0.9\columnwidth]{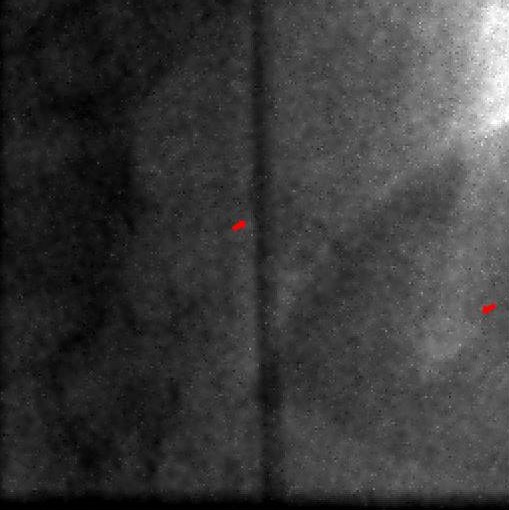}\label{fig:taba}\end{subfigure}&
			\begin{subfigure}{0.5\textwidth}\centering\includegraphics[width=0.9\columnwidth]{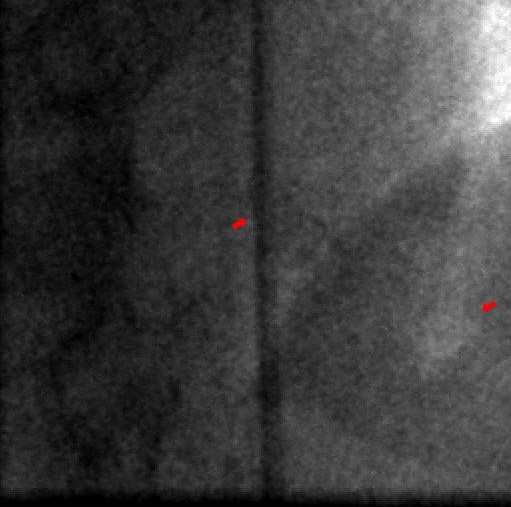}\label{fig:taba}\end{subfigure} \\\\
			
			\begin{subfigure}{0.5\textwidth}\centering\includegraphics[width=0.9\columnwidth]{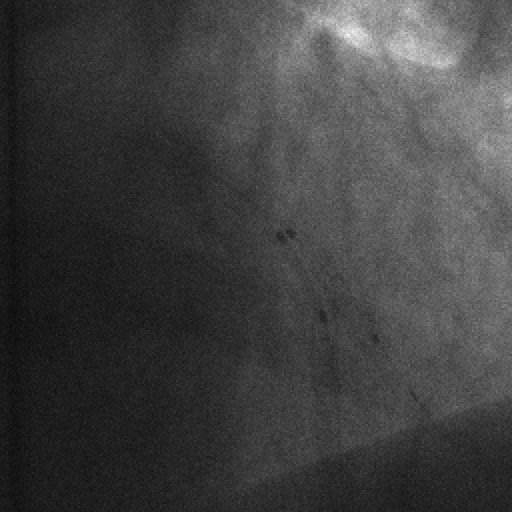}\label{fig:taba}\end{subfigure}&
			\begin{subfigure}{0.5\textwidth}\centering\includegraphics[width=0.9\columnwidth]{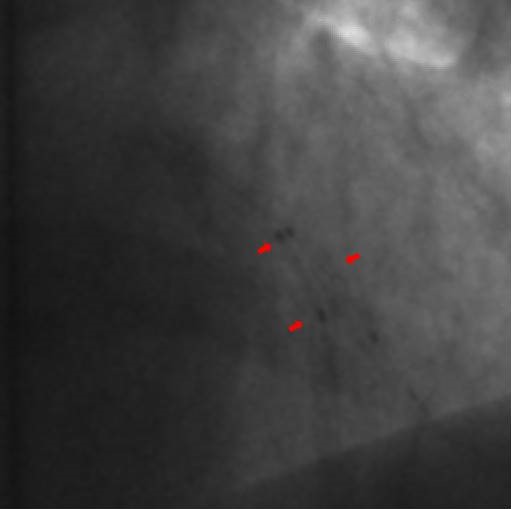}\label{fig:taba}\end{subfigure}&
			\begin{subfigure}{0.5\textwidth}\centering\includegraphics[width=0.9\columnwidth]{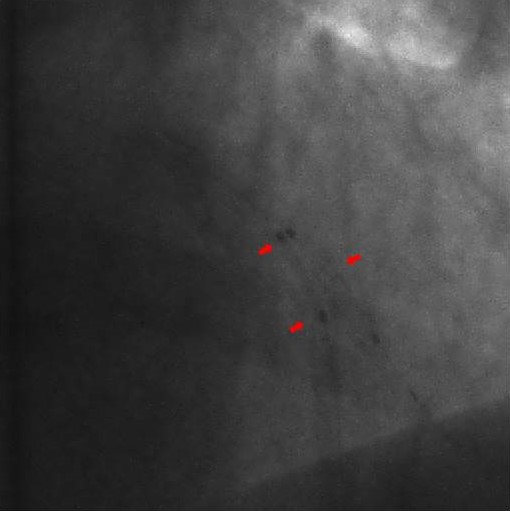}\label{fig:taba}\end{subfigure}&
			\begin{subfigure}{0.5\textwidth}\centering\includegraphics[width=0.9\columnwidth]{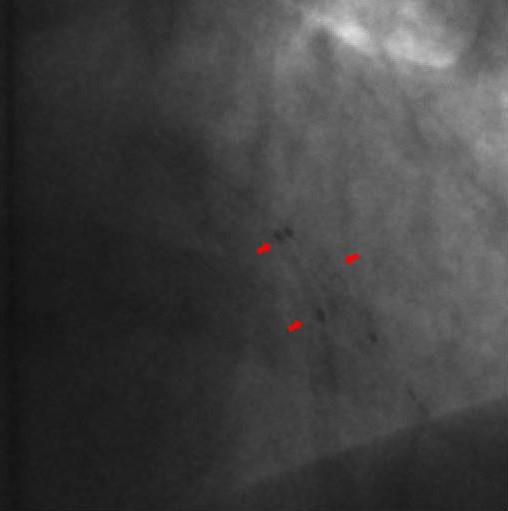}\label{fig:taba}\end{subfigure} \\\\
						
		\end{tabular}
	}
	\caption{Comparison of original low-dose image and its corresponding results obtained from different user losses for the first user. For better visualization windowing is applied on the second row.}
	\label{fig:firstuser}
\end{figure}

\begin{figure}[tb]
	\centering
	\graphicspath{{Plots/}}
	\resizebox{\columnwidth}{!}{
		\begin{tabular}{|cc|c|c|}
			\hline
			
			\rotatebox{90}{User 1} & \begin{subfigure}{0.5\textwidth}\centering\includegraphics[width=0.9\columnwidth]{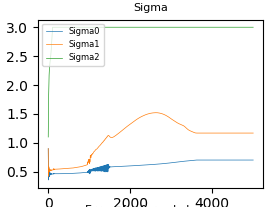}\caption{}\label{fig:taba}\end{subfigure}&
			\begin{subfigure}{0.5\textwidth}\centering\includegraphics[width=0.9\columnwidth]{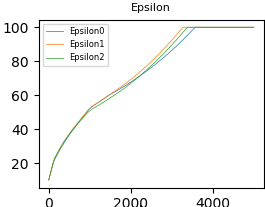}\caption{}\label{fig:taba}\end{subfigure}&
			\begin{subfigure}{0.5\textwidth}\centering\includegraphics[width=0.9\columnwidth]{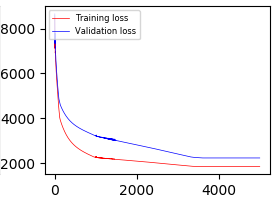}\caption{}\label{fig:taba}\end{subfigure}\\	
			
			\rotatebox{90}{User 2} & \begin{subfigure}{0.5\textwidth}\centering\includegraphics[width=0.9\columnwidth]{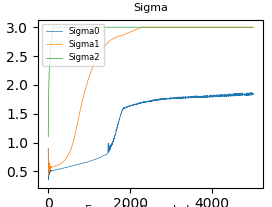}\caption{}\label{fig:taba}\end{subfigure}&
			\begin{subfigure}{0.5\textwidth}\centering\includegraphics[width=0.9\columnwidth]{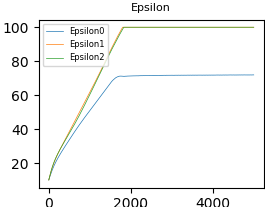}\caption{}\label{fig:taba}\end{subfigure}&
			\begin{subfigure}{0.5\textwidth}\centering\includegraphics[width=0.9\columnwidth]{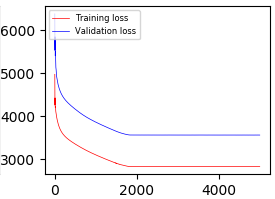}\caption{}\label{fig:taba}\end{subfigure}\\
            \hline
						
		\end{tabular}
	}
	\caption{Comparison of Hybrid loss for two users. Note that we set 100 as maximal value for $\epsilon$. User 1 favors sharper images while User 2 prefers smoother images.}
	\label{fig:usercomparision}
\end{figure}

Qualitative results of our approach are presented in Fig.~\ref{fig:firstuser} for the first user. These indicate an influence of different loss functions on the parameter tuning of one user's preferences. The \textit{Best Match} loss shows better noise reduction, however reduces the sharpness more than the other losses. In contrast to \textit{Best Match}, \textit{Forced Choice} loss shows better sharpness and higher noise level. In order to favor both targets the \textit{Hybrid Loss} eliminates noise and preserve sharpness of image data as well. 
Fig.~\ref{fig:usercomparision} displays the Hybrid loss curves for our two different users over the training process. It demonstrates that User 1 favors sharper images than User 1. Note that we set a value of 100 as maximum for parameters $\epsilon$.

\subsection{Quantitative Evaluation}
In this section, we evaluate the three loss functions for both of our users against each other. Table~\ref{tab:quant} displays the models created with the respective loss functions versus the test sets of both users. To set fair conditions for the comparision, we only evaluated models with the respective loss functions that were used in their training. The results indicate that Best-Match and Forced-Choice only are not able to result in the lowest loss for their respective user. The Hybrid loss models, however, are minimal on the test data of their respective user. Hence, the Hybrid loss seems to be a good choice to create user-dependent de-noising models. 
\begin{table}[tb]
	\caption{Quantitative comparison of loss functions: Best-Match (BM), Forced-Choice (FC), Hybrid(HY)}\label{tab:quant}
	\centering
		\begin{tabular}{|cc|ccc|ccc|}
				\hline

				\multirow{2}{*}{Low dose data}	& &\multicolumn{3}{c}{User 1}			& \multicolumn{3}{c|}{User 2}		\\\cline{3-8}
												& &BM 		& FC 		& HY 			& BM 		& FC 		& HY 		\\\hline
				\multirow{3}{*}{Model Nr. 1} 	& BM	&  ~1431.1 	&  	---		& 				&2436.7 	& 	---		&  ---		\\\cline{3-8}
												& FC &  ---			& ~248.8 	&  	&	---		& ~253.1 	&  ---\\\cline{3-8}
												& HY & 	---			&  	---		& ~{\bf 1771.1} &	---		& 	---	& 2675.9\\\hline
												
				\multirow{3}{*}{Model Nr. 1} 	& BM &  1381.5	&  	---		& 	---		& 2391.5 	& 	---		&  				\\\cline{3-8}
												& FC &  ---		& ~~249.5 	&  	---		&	---		& 964.9 	&  	---			\\\cline{3-8}
												& HY & 		---	&  	---		& ~1781.1 	&	---		& 	---		& ~{\bf 2359.1}	\\\hline																	
																			
		\end{tabular}
\end{table}

\section{Conclusion and Discussion}
We propose a novel user loss for neural network training in this work. It can be applied to any image grading problem in which users have difficulties in finding exact answers. As a first experiment for the user loss, we demonstrate that it can be used to train a de-noising algorithm towards a specific user. 
In our work 200 decisions using 50 clicks were sufficient to achieve proper parameter tuning. In order to be able to apply this for training, we used the \textit{precision learning} paradigm to create a suitable network structure with only few trainable parameters.

Obviously also other algorithms would be suited for the same approach  \cite{tomasi1998bilateral,petschnigg2004digital,Luisier,Motwani,c4}. However, as the scope of the paper is the introduction of the user loss, we omitted these experiments in the present work. Further investigations on which filter requires how many clicks for convergence is still an open question and subject of future work.

We believe that this paper introduces a powerful new concept that is applicable for many applications in image processing such as image fusion, segmentation, registration, reconstruction, and many other traditional image processing tasks.

\bibliographystyle{splncs04}
\bibliography{references}

\begin{thebibliography}{10}
\providecommand{\url}[1]{\texttt{#1}}
\providecommand{\urlprefix}{URL }
\providecommand{\doi}[1]{https://doi.org/#1}

\bibitem{Bishop:2006}
Bishop, C.M.: Pattern Recognition and Machine Learning (Information Science and
  Statistics). Springer-Verlag, Berlin, Heidelberg (2006)

\bibitem{vesselnet}
Fu, W., Breininger, K., Schaffert, R., Ravikumar, N., W{\"u}rfl, T., Fujimoto,
  J., Moult, E., Maier, A.: {Frangi-Net: A Neural Network Approach to Vessel
  Segmentation}. In: Maier, A., T., D., H., H., K., M.H., C., P., T., T. (eds.)
  {Bildverarbeitung f{\"u}r die Medizin 2018}. pp. 341--346 (2018)

\bibitem{lecun2015deep}
LeCun, Y., Bengio, Y., Hinton, G.: Deep learning. nature  \textbf{521}(7553),
  ~436 (2015)

\bibitem{Luisier}
Luisier, F., Blu, T., Unser, M.: A new sure approach to image denoising:
  Interscale orthonormal wavelet thresholding. IEEE Transactions on image
  processing  \textbf{16}(3),  593--606 (2007)

\bibitem{precision_learning}
Maier, A.K., Schebesch, F., Syben, C., W{\"{u}}rfl, T., Steidl, S., Choi, J.H.,
  Fahrig, R.: Precision learning: Towards use of known operators in neural
  networks. CoRR  \textbf{abs/1712.00374} (2017),
  \url{http://arxiv.org/abs/1712.00374}

\bibitem{Motwani}
Motwani, M.C., Gadiya, M.C., Motwani, R.C., Harris, F.C.: Survey of image
  denoising techniques. In: Proceedings of GSPX. pp. 27--30 (2004)

\bibitem{petschnigg2004digital}
Petschnigg, G., Szeliski, R., Agrawala, M., Cohen, M., Hoppe, H., Toyama, K.:
  Digital photography with flash and no-flash image pairs. In: ACM transactions
  on graphics (TOG). vol.~23, pp. 664--672. ACM (2004)

\bibitem{c4}
Rajashekar, U., Simoncelli, E.P.: Multiscale denoising of photographic images.
  In: The Essential Guide to Image Processing, pp. 241--261. Elsevier (2009)

\bibitem{rockafellar}
Rockafellar, R.: Convex Analysis. Princeton landmarks in mathematics and
  physics, Princeton University Press (1970),
  \url{https://books.google.de/books?id=1TiOka9bx3sC}

\bibitem{rumelhart1986learning}
Rumelhart, D.E., Hinton, G.E., Williams, R.J.: Learning representations by
  back-propagating errors. nature  \textbf{323}(6088), ~533 (1986)

\bibitem{filter_learning}
Syben, C., Stimpel, B., Breininger, K., W{\"u}rfl, T., Fahrig, R., D{\"o}rfler,
  A., Maier, A.: {Precision Learning: Reconstruction Filter Kernel
  Discretization}. In: Noo, F. (ed.) {Proceedings of the Fifth International
  Conference on Image Formation in X-Ray Computed Tomography}. pp. 386--390
  (2018)

\bibitem{tomasi1998bilateral}
Tomasi, C., Manduchi, R.: Bilateral filtering for gray and color images. In:
  Computer Vision, 1998. Sixth International Conference on. pp. 839--846. IEEE
  (1998)

\end{thebibliography}

\end{document}